\documentclass[letter]{ieice}
\usepackage{graphicx}
\usepackage{xcolor}
\usepackage[fleqn]{amsmath}
\usepackage{newtxtext}
\usepackage[varg]{newtxmath}
\usepackage{color}

\field{}

\title{RECOGNITION OF MULTIPLE FOOD ITEMS IN A SINGLE PHOTO FOR USE IN A BUFFET-STYLE RESTAURANT}
\authorlist{%
 \authorentry{Masashi Anzawa}{}{U-TOKYO}\MembershipNumber{}
 \authorentry{Sosuke Amano}{}{U-TOKYO}\MembershipNumber{}
 \authorentry{Yoko Yamakata}{}{U-TOKYO}\MembershipNumber{}
 \authorentry{Keiko Motonaga}{}{JISS}\MembershipNumber{}
 \authorentry{Akiko Kamei}{}{JISS}\MembershipNumber{}
 \authorentry{Kiyoharu Aizawa}{}{U-TOKYO}\MembershipNumber{}
}
\affiliate[U-TOKYO]{The University of Tokyo}
\affiliate[JISS]{Japan Institute of Sports Sciences}

\DeclareMathOperator*{\argmax}{arg\,max}
\DeclareMathOperator*{\onlyif}{only\,if}
\graphicspath{{./figure/}}

\begin{document}
\maketitle
\begin{summary}
We investigate image recognition of multiple food items in a single photo, focusing on a buffet restaurant application, where menu changes at every meal, and only a few images per class are available. After detecting food areas, we perform hierarchical recognition. We evaluate our results, comparing to two baseline methods. 
\end{summary}
\begin{keywords}
image recognition, food recognition, food recording tool
\end{keywords}

\section{Introduction}
\label{sec:intro}
\begin{figure}[!t]
	\centering
	\includegraphics[width=0.48\textwidth]{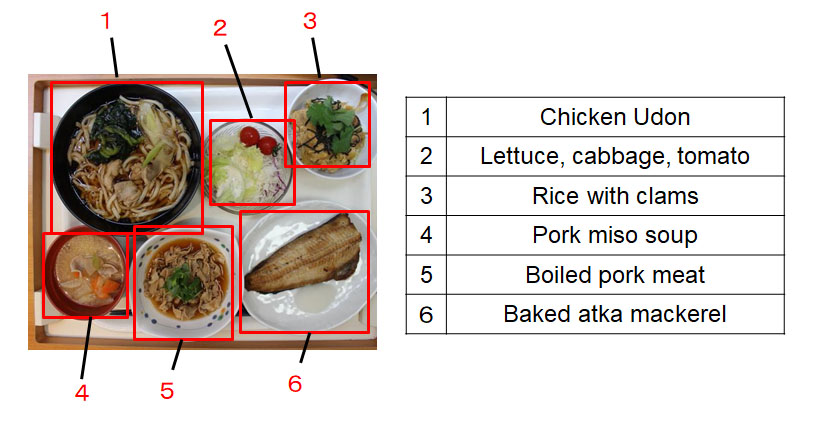}
	\caption{An example of the result of the food recognition}
	\label{img:concept}
\end{figure}

Nutrition management is very important in the health care industry \cite{Kim2016}. Nutritional intake is generally gauged via food records to help prevent lifestyle diseases (e.g., obesity and diabetes). A nutritionist, for example, promotes an individual's health by providing specific advice based on the person’s lifestyle information. Manually recording food intake is, however, a burden over the long term. In this paper, we focus on the application in a buffet-style restaurant. We investigate food recognition of multiple food items in a single photo, and we integrate food localization and hierarchical recognition. 
An example of recognition of a photo with its multiple food items is shown in Fig. \ref{img:concept}.

In a standard image recognition task, detectors are generally used with classifiers trained by large quantities of fixed-class datasets. It is difficult, however, to apply conventional strategies to a buffet-style restaurant. First, the menu changes every meal, and only one or a few template images per dish are available. Then, the amount of data is insufficient to fine-tune food classifiers. Second, buffet users freely take dishes on a tray, and some foods, such as vegetables, are mixed in a single dish.

This paper first localizes dishes and performs hierarchical recognition, during which, food is first recognized as a single class. Then, specific food classes are selected and detailed for localization and recognition. We obtain food images for our experiments from the Japan Institute of Sports Sciences (JISS). We evaluate accuracy results per tray and the error of the amounts of nutritional value.

Our contributions are summarized below.
\begin{itemize}
\item We create a processing pipeline consisting of food-region detection and recognition of multiple food items in a single photo, focusing on buffet use, where most food items change at every meal. The buffet restaurant data consists of one or a few close-up pictures of each food item.
\item We propose a hierarchical fine-grained recognition for specific categories of dishes, such as salads, to detect multiple items mixed on a plate. 
\item We collect photos from a real buffet restaurant over 10 days. Compared to baseline methods, which rely on single-class and multi-class recognition, the proposed method significantly improves performance.
\item We evaluate the recognition performance from the perspective of nutritional values (e.g., energy, protein, lipid, and carbohydrates) and verify that the error is significantly reduced by our method, compared to the baselines.  
\end{itemize}

\begin{figure*}[!t]
	\centering
	\includegraphics[width=0.95\textwidth]{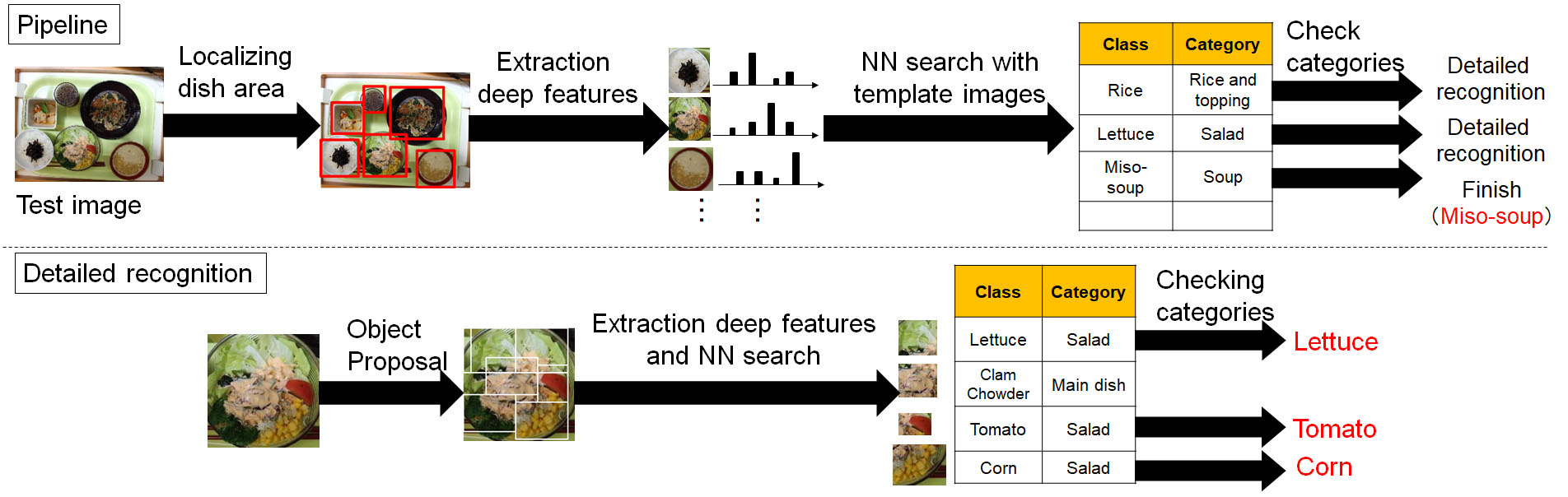}
	\caption{A pipeline of food localization and hierarchical recognition}
	\label{img:athlete_hierarchical_flow}
\end{figure*} 

\section{Related Work}
Food-image recognition using a fixed-class dataset has been very well studied \cite{Chen2009,Matsuda2012,Kawano2014,Bossard2014,kagaya2014,Bolanos2016,Martinel2016,He2016}. Many studies trained food classifiers in a convolutional neural network using a fixed-class large food image dataset. Martinel et al. \cite{Martinel2016} used a Wide-Slice ResNet and achieved 89.6\% for a 100-class dataset. Food image recognition accuracy in experimental environments is sufficiently high. However, it is not suitably high enough for our task, because the buffet restaurant menu changes at every meal, and only one or a few template images per dish are available.

Food image recognition using personal diet history has also been studied  
\cite{Aizawa2013,Aizawa2015}.  Aizawa et al.\cite{Aizawa2015} proposed a food retrieval system that searched input foods from the individual diet history using the nearest-neighbor (NN) search of image-feature vectors. Horiguchi et al. \cite{Horiguchi2018} investigated personalized food recognition using weighted NN recognition of common and personal food items. Yu et al. \cite{Yu2018} further extended personalized classifiers with weights optimized for food items and temporal changes. Whereas we do not consider the individuality of users in this paper, the NN classifier-based methodology, using a small number of templates, is similar.

Food recognition almost always uses datasets, but each photo contains only a single food item. For easier use of food recognition in daily life, it is desirable to recognize multiple food items in a single photo, where both food-region detection and food-item recognition can be applied. 

The applicability of food-image recognition to restaurants has been well-studied \cite{Bettadapura2015} \cite{Beijbom2015} \cite{Herranz2017} \cite{Meyers2015}. These methods used location information, such as Global Positioning System, to identify a target restaurant from a list of multiple restaurants. Then, they used food classifiers corresponding to the target restaurant. Consequently, their conditions were different from those of our task.

\begin{figure}[!t]
	\centering
	\includegraphics[width=0.48\textwidth]{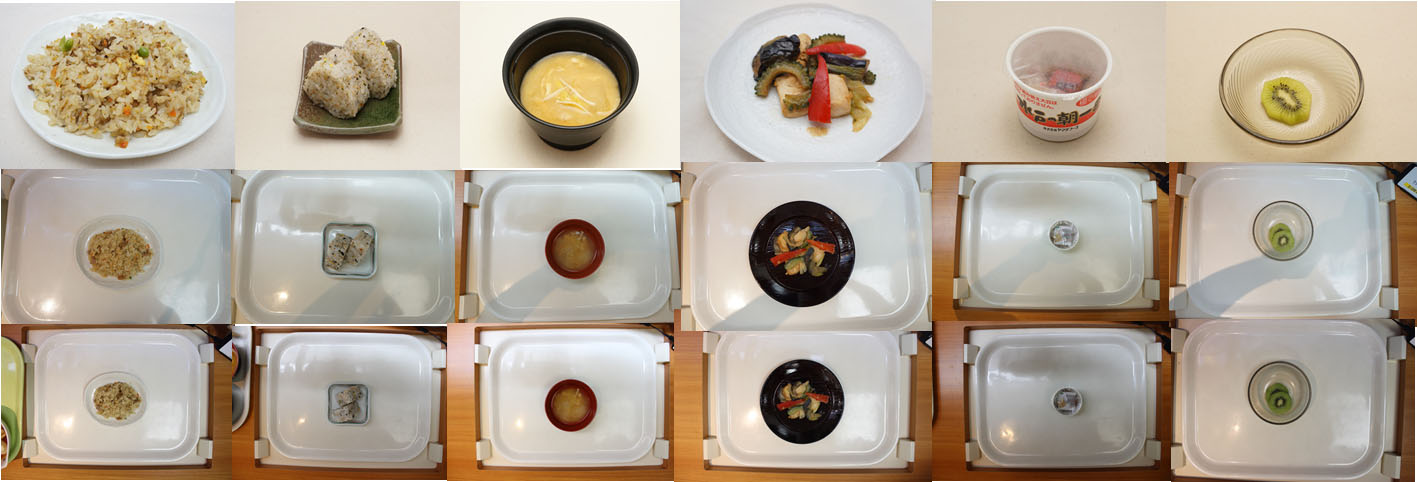}
	\caption{Examples of JISS-22 template images. The top is taken from a diagonal perspective and the other two are taken from directly above. Three photos exist for each food item.}
	\label{img:template}
\end{figure}

\begin{table}[!t]
	\centering
	\caption{A list of food items for one meal used in our experiments. Food items except "others" change at every meal. We exclude seasonings and dressings (e.g., vinegar, sesame oil), because they are not well-visible.}
	\begin{tabular}{|l|l|}	\hline
	      Category  &   Classes (food items) \\ \hline \hline
		Staple food & Beef bowl, Cold udon, Rice ball, Carbonara\\ 
		~~Rice and toppings & Rice, Salted plum, Chili oil, Kelp tsukudani \\ \hline 
		Main dish & Scrambled egg, Chicken Yawata roll \\
				& Mackerel of mirin marinated grilled\\ \hline
		Side Dish & Seasoned komatsuna, Boiled chiken, Oyaki \\
             ~~Salad & Cabbage, Lettuce, Tomato, Broccoli, Paprika\\
 			      & Shredded radish, Macaroni salad, Edamame \\
				& Seaweed konjak salad\\ \hline
             Soup stock	& Miso soup, Chinese corn soup \\ \hline
		Fruits & Pine, Pink grapefruit \\ \hline
		Dessert & Green tea roll cake, Salt lemon mousse cake \\ \hline 
		Others & Coffee, Iced coffee, Barley tea, Green tea, \\
			   & Processed soy milk, Natto, Egg, 6P cheese, Milk \\
                      &Low-fat milk, Plain yogurt, Honey, Grapefruit juice\\
                     & Orange juice, Tofu, Packaged furikake\\ \hline
	\end{tabular}
	\label{meal_example}
\end{table}

\section{Dataset} 
\label{sec:athlete_dataset}
In this paper, we use two datasets: JISS for experiments and FoodLog App for training the detector and classifier.

\subsection{JISS Dataset}
\label{sub:athlete_dataset_jiss}
We use real data obtained from the restaurant in JISS\footnote{https://www.jpnsport.go.jp/jiss/}. We refer to this dataset as JISS-22. It shows 22 meal-data provided from August 1st through 10th, 2017 (i.e., 8 breakfasts, 7 lunches, and 7 dinners). About 50 items (dishes and foods) are provided at each meal. Most of items change at every meal. A list of dishes for one meal is shown in Table \ref{meal_example}.

An item's name is a class ($y$) to be recognized. A category is a set of related classes (e.g., salad category ($c$)) that contain classes of lettuce, tomato, etc. Each food item is associated with nutritional information, and nutrition taken by an individual per meal is computable by estimating its classes. In this paper, we use energy and three major nutrients (e.g., protein, lipids, carbohydrates). Three template images for each class are available. Examples of template images are shown in Fig. \ref{img:template}. 

A user places various dishes on a tray from approximately 50 available items, and multiple dishes appear in each test image. Additionally, a salad dish often contains multiple classes, such as lettuce and tomato. The number of total test images is 195. Examples of JISS-22 test images are shown in Fig. \ref{img:concept}, where multiple dishes are placed on a tray.

\begin{figure}[!t]
	\centering
	\includegraphics[width=0.48\textwidth]{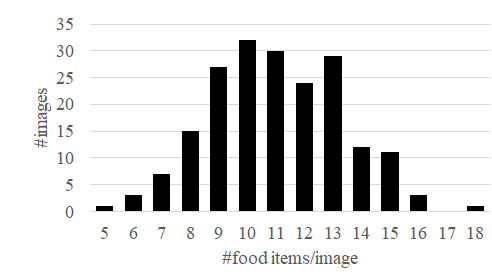}
	\caption{Frequency of number of food items per photo of JISS-22.}
	\label{JISS22-freq}
\end{figure}

JISS-22 only has annotations of food items for each photo and does not have bounding boxes of dishes.
The average number of food items per photo is 11.0 and the standard deviation is 2.3. The frequency of 
the number of food items per photo of JISS-22 is shown in Fig. \ref{JISS22-freq}.
The number of food items does not equal to the number of dish plates because multiple food items are 
mixed in one plate such as salad.   

\begin{figure}
	\centering
	\includegraphics[width=0.48\textwidth]{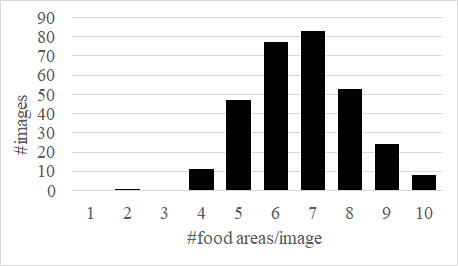}
	\caption{Frequency of number of dish bounding boxes per photo of JISS-DET. }
	\label{JISS-DET-freq}
\end{figure}

Additionally, we use a different set of photos of the JISS buffet restaurant data 
taken during non-overlapping period of time
to fine-tune the dish area detector (i.e., JISS-DET). 
JISS-DET consists of 304 images with
bounding box annotations of dish areas. 
The average number of bounding boxes of dish areas is 6.7 and its standard deviation is 1.41. 
The frequency of number of bounding boxes per image of JISS-DET is shown in Fig. \ref{JISS-DET-freq}.

\subsection{FoodLog Dataset}
\label{sub:athlete_dataset_fl}
The JISS dataset is not large enough to train the detectors and classifiers. Therefore, we use the FoodLog Dataset (FLD), built with FoodLog App \cite {Aizawa2015} for training. We use data consisting of 450,066 images with rectangular annotation to train the dish area detector (i.e., FLD-DET). 

We also use FLD-469, consisting of 234,500 images from FLD, representing 469 classes, each having 500 images for the classifier training. All images are resized to 256 x 256. We train the network with 469 classes, and we use deep features obtained from the last pooling layer.  

\section{Proposed Method}
We propose a framework of automatic food recognition for multiple food items in a single photo. We first localize dishes. Then, we perform hierarchical recognition, during which, food is recognized as a single class. Then, specific food classes are selected, and further detailed localization and recognition are performed. The entire processing pipeline is shown in Fig. \ref{img:athlete_hierarchical_flow}. 

Because the menu changes at every meal, the template images are updated at each meal. Instead of training classifiers for every meal, we use deep features and apply NN searches on the template and target features. We take deep features, $x_{i}$, from the output of the last average pooling layer of ResNet 50, to which we apply L2-normalization. The NN of deep features works well for classification tasks when enough data is used for training the network \cite{Horiguchi2018,Horiguchi2017}. 

The template images, $V_{m}$, are denoted by
\begin{equation}
\label{eq:athlete_v}
V_{m}=\{(x_{i}, y_{i})|1 \leq i \leq N_{m}\},
\end{equation}
where $y_ {i}$ and $x_{i}$ represent the class and feature vectors, respectively. $N_{m}$ is the total number of template images contained in the meal.

As shown in Fig. \ref{img:athlete_hierarchical_flow}, we input the test image and obtain dish-area candidates, $B_{i}(1\leq i \leq N_{B})$, by the dish area detector, where $N_{B}$ is the total number of dish-area candidates. Then, we extract deep features, $x_{Bi}$, for $ B_{i}$. The class, $y_{B_{i}}^{\ast}$, is estimated by the NN search, denoted by
\begin{equation}
\label{eq:athlete_ybi}
y_{B_{i}}^{\ast} = \argmax_{y \in Y} \{s(y, x_{B_{i}}, V_{m}) \},
\end{equation}
where $Y$ is a set of classes included in $V_ {m}$. The similarity, $s(\cdot)$, an inner product, is denoted by
\begin{equation}
\label{eq:athlete_s}
s(y, x_{B_{i}}, V_{m}) = \underset{x \in V_{y}}{\max} x^{T}x_{B_{i}},
\end{equation}
where $V_ {y}$ represents vectors of the class, $y$. In our case,  $V_ {y}$ contains at most three template vectors for each class, $y$.

The NN-based classification of features works for the condition of a small number of template images. We use it for a single-class recognition method. If the dish area contains only one food item, this single-class recognition is satisfactory. However, a food item, such as a vegetable, is almost always accompanied by several others in a salad dish.

We choose food items corresponding to specific categories (e.g., "salad", "fruits", "rice and toppings") for the fine-detailed recognition in the dish area. 
Example of the three categories are shown in Table \ref{meal_example}. 
When the result of the single class recognition is in these specific categories - e.g. if the result
of the single class recognition of a region is tomato, which is in salad category, the region is performed 
fine-grained recognition. Regions classified to food items included the three categories are further analyzed by fine-grained recognition. 

The fine-detailed recognition is performed as follows.
We generate object candidates, $B_ {i, j}$, within the area of $ B_ {i}$. In the experiments, we simply apply a sliding window to produce candidates. Then, we perform single-class recognition on each object candidate region, $B_ {i, j}$, and estimate the class, $y_{B_{i,j}} ^ {\ast}$. The estimation, $Y_{B_{i,j}} ^ {\ast}$, is accepted if the category of the class is the same as that of $B_ {i}$. 

\begin{equation}
\label{eq:athlete_Ybi_detail}
Y_{B_{i}}^{\ast} = \{y_{B_{i,j}}^{\ast}  \onlyif  c_{B_{i,j}}^{\ast} = c_{B_{i}}^{\ast},1 \leq j \leq N_{B'} \}.
\end{equation}
In other words, in fine-grained recognition, we exclude classes whose categories are different from that estimated in the single-class recognition. 

\begin{table}[!t]
	\centering
	\caption{The results of food image recognition and nutrient estimation for JISS-22.}
	\begin{tabular}{|l|c|c|c|}
		\hline
		&   Single-class  &   Multi-class    & \textbf{Hierarchical} \\
		&   recognition &   recognition   &  \textbf{recognition} \\ \hline \hline
		Precision & \textbf{0.881} & 0.741 & 0.821 \\ \hline
		Recall & 0.591 & 0.722 & \textbf{0.755}  \\ \hline
		F-measure & 0.708 & 0.731 & \textbf{0.787} \\ \hline
		MAE Energy (kcal) & 106 (13.8\%) &152 (20.2\%) & \textbf{74 (9.4\%)} \\ \hline
	      MAE Protein (g) & 4.93 (11. 9\%) & 8.76 (21.0\%) & \textbf{4.04 (9.6\%)} \\ \hline
		MAE Lipid (g) & 4.27 (16.3\%) & 6.77 (24.8\%) & \textbf{3.72 (14.3\%)} \\ \hline
		MAE Carbonhydrate (g) & 14.0 (16.2\%) & 18.1 (22.6\%) & \textbf{8.98 (9.9\%)} \\ \hline
	\end{tabular}
	\label{tab:athlete_result}
\end{table}

\section{Experiment} 
\label{sec:athlete_exp}
\subsection{Implementation and Evaluation Metric}
\label{sub:athlete_expe_eval}
Regarding dish-area detection, we utilized SSD300\cite{Liu2016}, trained using FLD-DET and fine-tuned with JISS-DET. Regarding feature extraction, we utilized pre-trained ResNet50 \cite{ResNet}, which we fine-tuned with FLD-469. 

To evaluate of the proposed method, we compared ours with two other baseline methods. One used single-class recognition, which is equivalent to the estimations of the first step of hierarchical recognition. The other was multi-class recognition, for which we calculated the similarity of classes in a dish area and estimated multiple classes by thresholding their similarity. The dish area detection method (i.e., SSD300) was the same for the proposal and the comparison methods. 

Test data was JISS-22. For multi-class recognition, JISS-22 was divided into 1/3 and 2/3. We used 1/3 for optimization of threshold and 2/3 for testing. We experimented with cross validation.

The metrics used for evaluation were precision, recall, and F-measure of recognition and the mean absolute error (MAE) of nutritional information. 

All the buffet dishes have their nutrition information. Assuming the amount of the dish is one serving size 
and summing all the dishes on a tray, we computed the total nutrition of the tray. 
We compared energy (kcal) between those of the recognized dishes and GT dishes. 
It shows influence of recognition accuracy from the point of view of energy (kcal) 
without considering variation of the amount. It may change when the serving size is 
changed - some of food items such as rise are self-served, and the real value can
change.   

\begin{figure}[!t]
	\centering
	\includegraphics[width=0.45\textwidth]{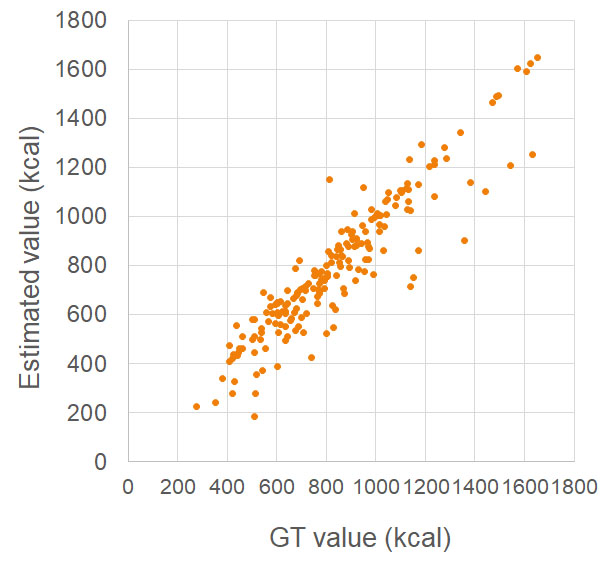}
	\caption{Distribution of GT values and estimated values of energy (kcal) in hierarchical recognition (sliding window).}
	\label{img:athlete_calorie}
\end{figure}

\begin{figure}[!t]
	\centering
	\includegraphics[width=0.48\textwidth]{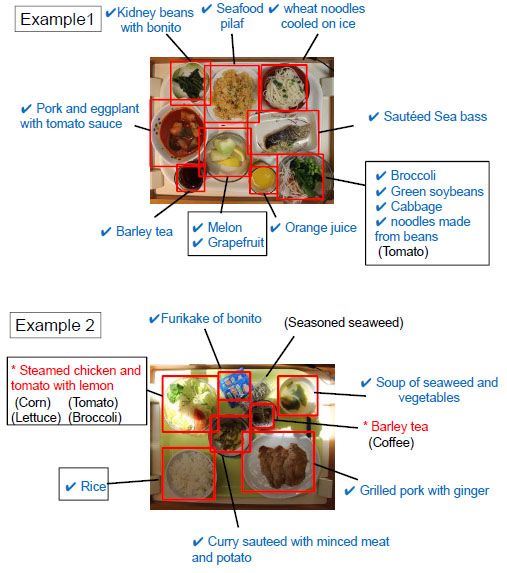}
	\caption{Examples of results by hierarchical recognition. Blue letters with check marks correspond to correct answers, and red letters with star marks correspond to incorrect answers. Black letters with parentheses are missing. The rectangles is the region of detailed recognition. }
	\label{img:athlete_hierarchical_example}
\end{figure}

\subsection{Result}
\label{sub:athlete_expe_result}
Table \ref{tab:athlete_result} shows the results of food-image recognition and nutrient estimation for JISS-22. Experimental results show that our approach achieved 0.79 as an F-measure and 74 kcal (9.4\%) error in energy: significantly better than baseline methods without the hierarchical scheme. 

Figure \ref{img:athlete_calorie} shows the distribution of GT values and the estimated values of energy (kcal) via hierarchical recognition. The correlation coefficient between the GT values and the estimated values was 0.92, which is sufficiently high.

Finally, we show the examples of results by hierarchical recognition in Fig. \ref{img:athlete_hierarchical_example}. In Example 1, we see that multiple classes within the “salad” category on one dish were correctly estimated except "tomato" which is not very visible . We can see similar results with “fruit.”Example 2 shows a failure case, where detailed recognition of the salad dish did not work, because vegetable was not estimated by single-class recognition.

\section{Conclusion}
In this paper, we proposed a framework of automatic food recognition, with a focus on a buffet-style restaurant. In our processing pipeline, we first localized dishes and performed hierarchical recognition, during which, food was recognized as a single class. Then, specific food classes were selected and further detailed. Additionally, localization and recognition were performed. We obtained food images from the JISS dataset for our experiments. We evaluated accuracy results per tray and the error in the amount of nutritional value. Experimental results using real data showed that our approach can achieve 0.79 in F-measure and 9.4\% error in energy, significantly better than the baseline methods without the hierarchical scheme. 

\subsubsection*{ACKNOWLEDGEMENTS}
This work was partially supported by JST CREST JPMJCR1686 and JSPS KAKENHI 18H03254.


\begin{thebibliography}{1}

\bibitem{Kim2016}
Y. Kim, S. Ji, H. Lee, J.W. Kim, S. Yoo, and J. Lee, “My doctor is keeping an eye on me!: Exploring the clinical applicability of a mobile food logger,” ACM CHI2016, pp.5620-5631.

\bibitem{Chen2009}
M. Chen, K. Dhingra, W. Wu, L. Yang, R. Sukthankar, and J. Yang, “Pfid: Pittsburgh fast-food image dataset,” IEEE ICIP2009, pp.289-292.

\bibitem{Matsuda2012}
Y. Matsuda, H. Hoashi, and K. Yanai, “Recognition of multiple-food images by detecting candidate regions,” IEEE ICME2012, pp.25–30.

\bibitem{Kawano2014}
Y. Kawano and K. Yanai, “Automatic expansion of a food image dataset leveraging existing categories with domain adaptation,” ECCV2014 Workshop, pp.3-17.

\bibitem{Bossard2014}
L. Bossard, M. Guillaumin, and L. Van Gool, “Food-101–mining discriminative components with random forests,” ECCV2014, pp.446-461.

\bibitem{kagaya2014}
H. Kagaya, K. Aizawa, and M. Ogawa, “Food detection and recognition using convolutional neural network,” ACM Multimedia2014, pp.1085-1088.

\bibitem{Bolanos2016}
M. Bolanos and P. Radeva, “Simultaneous food localization and recognition,” ICPR2016, pp.3140-3145.

\bibitem{Martinel2016}
N. Martinel, G.L. Foresti, and C. Micheloni, “Wide-slice residual networks for food recognition,” arXiv:1612.06543, 2016.


\bibitem{He2016}
H. He, F. Kong, and J. Tan, “Dietcam: Multiview food recognition using a multikernel svm,” IEEE Journal of Biomedical and Health Informatics, vol.20, no.3, pp.848-855, 2016.

\bibitem{Aizawa2013}
K. Aizawa, H. Maruyama, Li, and C. Morikawa, “Food balance estimation by using personal dietary tendencies in a multimedia food log,” IEEE Trans. Multimedia, vol.15, no.8, pp.2176-2185, 2013.

\bibitem{Aizawa2015}
K. Aizawa and M. Ogawa, “Foodlog: Multimedia tool for healthcare applications,” IEEE MultiMedia, vol.22, no.2, pp.4-8, 2015.

\bibitem{Horiguchi2018}
S. Horiguchi, S. Amano, M. Ogawa, and K. Aizawa, “Personalized classifier for food image recognition,” IEEE Trans. Multimedia, vol.20, no.10, pp.2836-2848, 2018.

\bibitem{Yu2018}
Q. Yu, M. Anzawa, M. Ogawa, and K. Aizawa, “Food image recognition by personalized classifiers,” IEEE ICIP2018, pp.171–175.

\bibitem{Bettadapura2015} 
V. Bettadapura, E. Thomaz, A. Parnami, G.D. Abowd, and I. Essa, “Leveraging context to support automated food recognition in restaurants,” IEEE WACV2015, pp.580–587.

\bibitem{Beijbom2015} 
O. Beijbom, N. Joshi, D. Morris, S. Saponas, and S. Khullar, “Menu-match: Restaurant-specific food logging from images,” IEEE WACV2015, pp.844-851.

\bibitem{Herranz2017} 
L. Herranz, S. Jiang, and R. Xu, “Modeling restaurant context for food recognition,” IEEE Transactions on Multimedia, vol.19, no.2, pp.430-440, 2017.

\bibitem{Meyers2015}
A. Meyers, N. Johnston, V. Rathod, A. Korattikara, A. Gorban, N. Silberman, S. Guadarrama, G. Papandreou, J. Huang, and K.P. Murphy, “Im2calories: towards an automated mobile vision food diary,”
ICCV2015, pp.1233-1241.

\bibitem{Horiguchi2017}
S. Horiguchi, D. Ikami, and K. Aizawa, “Significance of softmax-based features in comparison to distance metric learning-based features,” arXiv:1712.10151, 2017.

\bibitem{Liu2016}
W. Liu, D. Anguelov, D. Erhan, C. Szegedy, S. Reed, C.Y. Fu, and A.C. Berg, “Ssd: Single shot multibox detector,” ECCV2016, pp.21-37.

\bibitem{ResNet}
K. He, X. Zhang, S. Ren, and J. Sun, “Deep residual learning for image recognition,” CVPR2016, pp.770-778.

\end{thebibliography}

\end{document}